\documentclass[sigconf,authorversion,nonacm]{acmart}

\usepackage[title]{appendix}
\usepackage{multirow} 
\usepackage{xcolor}
\usepackage{amsmath}
\usepackage{amsfonts}
\usepackage{overpic}
\usepackage[thinc]{esdiff}
\usepackage{array} 
\usepackage{caption} 
\usepackage{bbm}
\usepackage{float}

\newcommand{\ourmethod}{GAM}
\newcommand{\gc}{GC}
\newcommand{\gcp}{GC++}

\usepackage[switch]{lineno}  %
\usepackage{subfiles}
\begin{document}

\title{GAM: Explainable Visual Similarity and Classification via Gradient Activation Maps}


\author{Oren Barkan}
\authornote{Authors contributed equally to this work.}
\affiliation{%
  \institution{The Open University \& Microsoft}
  \country{Israel}
}
\author{Omri Armstrong}
\authornotemark[1]
\affiliation{%
  \institution{Tel-Aviv University}
  \country{Israel}
}
\author{Amir Hertz}
\authornotemark[1]
\affiliation{%
  \institution{Microsoft}
  \country{Israel}
}
\author{Avi Caciularu}
\affiliation{%
  \institution{Bar-Ilan University}
  \country{Israel}
}
\author{Ori Katz}
\affiliation{%
  \institution{Technion \& Microsoft}
  \country{Israel}
}

\author{Itzik Malkiel}
\affiliation{%
  \institution{Microsoft \& Tel Aviv University}
  \country{Israel}
}

\author{Noam Koenigstein}
\affiliation{%
  \institution{Microsoft \& Tel-Aviv University}
   \country{Israel}
}
\begin{abstract}
  We present Gradient Activation Maps (GAM) - a machinery for explaining predictions made by visual similarity and classification models.
By gleaning localized gradient and activation information from multiple network layers, GAM offers improved visual explanations, when compared to existing alternatives. The algorithmic advantages of GAM are explained in detail, and validated empirically, where it is shown that GAM outperforms its alternatives across various tasks and datasets.
\end{abstract}






\maketitle

\section{Introduction}
\label{sec:intro}

As the AI revolution disrupts industries and penetrates all walks of life, a growing need arises to intuitively explain machine-based decisions \cite{Vellido12makingmachine,Accountability_of_AI}. As a result, an emerging research area revolves around the need to make machine learning models more \emph{explainable}. 
This work joins this common effort and presents Gradient Activation Maps (GAM) - a novel method for explaining visual similarity and classification networks.

\begin{figure}[t]
\label{fig:intro-example}
    \centering
    \includegraphics[width=\columnwidth]{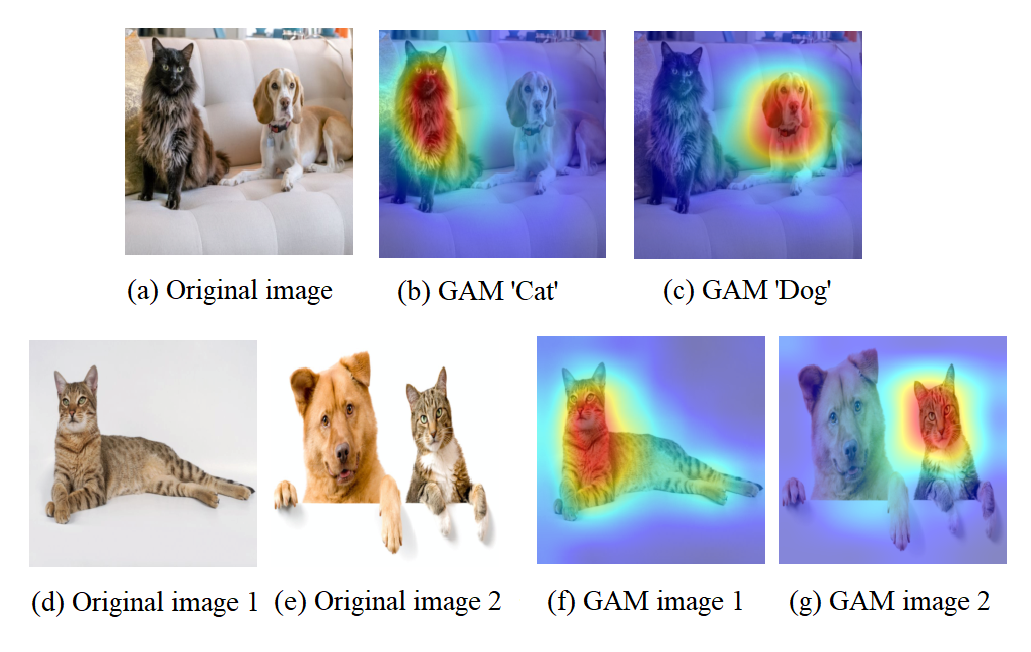}
    \caption{Visual explanations produced by GAM for similarity and classification tasks (using a pretrained DenseNet201). (b-c) w.r.t. the scores for 'cat' and 'dog' classes. (f-g) w.r.t. the cosine similarity between the latent representations of the images in (d-e).}
    \label{fig:intro_exmples}
\end{figure}

A \emph{saliency} map is an image depicting the relative contribution of each pixel in the input image w.r.t. the model's prediction. For example, Fig.~\ref{fig:intro_exmples} presents saliency maps produced by GAM for a classification task (a-c) and a similarity task (d-e). 
According to \cite{selvaraju2017grad}, a ‘good’ visual explanation technique should be (1) class discriminative i.e., localize the object in the image, and (2) high-resolution i.e., capture fine-grained details. 
However, comparing different visual explanation approaches is hard: A real methodological challenge stems from the lack of a ground-truth or a principled evaluation procedure. 
Hence, different works employed different evaluation procedures, often resorting to subjective visual assessments \cite{simonyan2013deep, stylianou2019visualizing}.

An actionable testing procedures for assessing the validity of saliency maps were recently proposed by Adebayo \cite{adebayo2018sanity}. 
Their work revealed that despite producing quality looking visualizations, most state-of-the-art methods produce saliency maps that are independent of either the model or the input-label relation, rendering them inadequate for producing explanations. 
An exception was the Grad-CAM (GC) method from \cite{selvaraju2017grad} that stood out among all others in its ability to produce fine-grained saliency maps, while successfully passing all sanity tests \cite{adebayo2018sanity}. Following the success of GC, an improved extension called Grad-CAM++ (GC++) was introduced and shown to outperform its predecessor on various visual explanation tasks \cite{chattopadhay2018grad}. 

GAM poses significant improvements upon GC and GC++ via several algoritmic features: Gradient localization, multi-layer analysis, and negative gradients suppression. These unique features lead to better saliency maps in terms of resolution, class discrimination, and object localization.
In Sec.~\ref{subsec:gam-vs-gc}, we elaborate on the relation of GAM to GC and GC++ and explain the algorithmic advantages behind GAM's superior results.

Our contributions are as follows:
\begin{itemize}
    \item We introduce GAM - a state-of-the-art method for extracting accurate saliency maps in terms of resolution and class discrimination. GAM is shown to outperform its alternatives on various objective and subjective evaluations, across all metrics, and especially in the case of small objects.
    \item We present a unified formulation for visual similarity and classification that enables the utilization of GAM, GC and GC++ for explaining visual similarity models (a task that was overlooked in \cite{selvaraju2017grad,chattopadhay2018grad}).
    \item We identify and demonstrate the limitations of GC and GC++, and explain how GAM averts these problems.
\end{itemize}


\section{Related work}
\label{sec:related}

\subsection{Explaining Visual Classification Models}

The early methods proposed by \cite{zeiler2011adaptive, zeiler2014visualizing,springenberg2014striving,simonyan2013deep, yosinski2015understanding,mahendran2016visualizing,yu2014visualizing,gradsam} are seminal works in visualization and understanding deep NNs.
Guided Backpropagation (GBP)  ~\cite{springenberg2014striving} visualizes the output prediction by propagating the gradients through the model and suppressing all negative gradients along the backward pass. However, GBP was shown to produce saliency maps that are not class discriminative~\cite{selvaraju2017grad}.
Another approach \cite{simonyan2013deep}, uses the gradients of predicted class scores w.r.t. to the input image to generate saliency maps.

Recently, Grad-CAM (GC)~\cite{selvaraju2017grad} created saliency maps based on the activations and gradients from the last convolution layer. In GC, the gradients of each channel are pooled to scalars. Then, these scalars weigh their corresponding activation maps that are summed together to produce the final saliency map. More recently, Grad-CAM++ (GC++)~\cite{chattopadhay2018grad} was introduced as an improved version of GC. GC++ uses a weighted average of the pixel-wise gradients in order to create the weights for the activation maps.

Both GC and GC++ operate on the last convolutional layer and employ gradient pooling that leads to the loss of gradient localization. In contrast, GAM utilizes the raw gradients from multiple layers in the network, enabling gradient localization with improved resolution and class discrimination. 

\subsection{Explaining Visual Similarity Models}
Previous works attempted to visually explain the decision made by similarity networks \cite{yi2014deep, schroff2015facenet,sun2014deep,wang2014learning,hoffer2015deep,oh2016deep}. These  networks are optimized to cluster images that are considered \textit{similar}, in a learned vector space. 
Other methods~\cite{radenovic2016cnn, tolias2015particular} determined areas that contributed to image similarity by comparing filter responses of images patches.
In \cite{chen2020adapting}, the authors utilized GC for explaining embedding networks that were trained on similarity tasks. However, their method is independent of the similarity score itself, hence it cannot be considered a ``true'' explanation to similarity. 

Recently, VDSN \cite{stylianou2019visualizing} was introduced as a method for visual explanation for similarity networks. VDSN produces saliency maps for image-pairs by combining the activations of the last convolution layer before and after average / max pooling. However, unlike GAM that utilizes the gradients of the similarity w.r.t. the activations from multiple layers, VDSN does not use the gradients, and hence is indepedent of the similarity score. Moreover, VDSN is limited to use the last convolutional layer in architectures that employ average / max pooling, and is applicable to similarity networks only (thus unable to visually explain classification models). 


\vspace{-2mm}
\subsection{Evaluating Saliency Maps}
\label{subsec:related-sanity}
Evaluating saliency maps is challenging, as no real ``ground truth'' exists, and the quality of an explanation is often subjective. 
In \cite{simonyan2013deep, selvaraju2017grad}, evaluations conducted using a weakly supervised object localization task, where the output saliency map is being used to specify the region in the image in which the classified object appears.
We further extend this approach to the image similarity task, by using the saliency maps to specify the regions in which similar objects appear in both images.

In~\cite{chattopadhay2018grad} the authors suggested the \emph{Average Drop Percentage (ADP)} and the \emph{Percentage of Increase in Confidence (PIC)} metrics, to measure the change in the model confidence when using \emph{explanation maps} (Hadamard product of the saliency map with the original image) instead of original image. We follow these tests and further extend them to the image similarity task.

In \cite{adebayo2018sanity}, the authors suggest sanity tests for saliency maps methods: The \emph{parameter randomization} and \emph{data randomization} procedures test whether the produced saliency map is sensitive to the randomization of the model's parameter and data labels, respectively. Otherwise, the method fails to faithfully explain the model's prediction.
Despite producing quality looking visualizations, the tests from \cite{adebayo2018sanity} reveal that many of the popular saliency methods do not pass the tests, and therefore are not adequate for providing satisfactory model explanations. In Appendix~\ref{sec:res_sanity} we show that GAM passes these tests.

\section{Gradient Activation Maps (GAM)}
\label{sec:method}

\subsection{A Unified Formulation for Visual Similarity and Classification}
\label{subsec: notations}

We begin by defining notations for the network's input and (internal) building blocks. The network's input is an image, denoted by ${x\in\mathbb{R}^{c_0 \times u_0 \times v_0}}$. The 3D activation produced by the $l$-th convolutional layer (for the image $x$) is denoted by ${h^l_x \in\mathbb{R}^{c_l \times u_l \times v_l}}$, where ${1 \leq l \leq L}$. Note that $h^l_x$ is not necessarily produced by a plain convolutional layer, but can be the output of a more complex function such as a residual \cite{he2016deep} or DenseNet \cite{huang2017densely} block. We further denote ${h^{lk}_x \triangleq h^l_x[k] (\in \mathbb{R}^{u_l \times v_l})}$ as the $k$-th activation map in $h^l_x$.

Let ${f:\mathbb{R}^{c_L \times u_L \times v_L} \rightarrow \mathbb{R}^d}$ be a function that maps 3D tensors to a $d$-dimensional vector representation. We denote the mapping of the last activation maps ${h^L_x}$ by ${f_x \triangleq f(h^L_x) \in \mathbb{R}^d}$. Note that $f$ may vary between different network architectures. Usually, it consists of a (channel-wise) global average pooling layer that is optionally followed by subsequent fully connected (FC) layers.

Finally, let ${s:\mathbb{R}^d \times \mathbb{R}^d \rightarrow \mathbb{R}}$ be a scoring function that receives two vectors and outputs a score. The use of $s$ varies between tasks: classification and similarity. In classification tasks, $f$ represents the last hidden layer of the network. The logit score for the class $j$ is computed by ${s(f_x,w_j)}$, where ${w_j \in \mathbb{R}^d}$ is the weights vector associated with the class $j$. In multiclass (multilabel) classification, $s$ is usually set to the dot-product, optionally with bias correction, or the cosine similarity. Then, either a softmax (sigmoid) function, with some temperature, transfers $s$ values to the range $[0,1]$.

For similarity tasks, we consider two images ${x,y\in\mathbb{R}^{c_0 \times u_0 \times v_0}}$, and a similarity score: $s(f_x,f_y)$. A common practice is to set $s$ to the dot-product or cosine similarity. 
Further note that in the specific case of similarity, the representation produced by $f$ is not necessarily taken from the last hidden layer of the network. Therefore, $f$ can be set to the output from any FC layer. For the sake of brevity, from here onward, we abbreviate both ${s(f_x,w_j)}$ and ${s(f_x, f_y)}$ with $s$. Disambiguation will be clear from the context.

\subsection{The GAM Method}
Given an image $x$, we denote the $l$-th saliency map ${m_x^l \in \mathbb{R}^{u_0 \times v_0}}$ by:
$
    m_x^l \triangleq m(h_x^l, g_x^l),
$
which is a function of the activation maps $h_x^l$ and their gradients: $g_x^l \triangleq \frac{\partial s}{\partial {h_x^l}}$. 
We denote $g^{lk}_x \triangleq g^l_x[k]$ (similarly to the notation $h^{lk}_x$). 
Then, we implement $m(h_x^l, g_x^l)$ as:
\begin{equation}
\label{eq:l-heatmap}
   m(h_x^l, g_x^l) = \textrm{NRM} \left[\textrm{RSZ}\left[\sum_{k=1}^{c_l}\phi(h_x^{lk}) \circ \phi(g_x^{lk})\right]\right],
\end{equation}
where $\phi$ is the ReLU activation function, and $\circ$ is the Hadamard product.
RSZ denotes the operation of resizing to a matrix of size $u_0 \times v_0$ (the height and width of the original image $x$). NRM denotes the min-max normalization.

The motivation behind Eq.~\ref{eq:l-heatmap} is as follows: each filter $k$ in the $l$-th convolutional layer captures a specific pattern. Therefore, we expect $h^{lk}_x$ to have high (low) values in regions that do (not) correlate with the $k$-th filter. In addition, regions in $g^{lk}_x$ that receive positive (negative) values indicate that increasing the value of the same regions in $h^{lk}_x$ will increase (decrease) $s$ value.

GAM highlights pixels that are both positively activated and associated with positive gradients. 
To this end, we first truncate all negative gradients (using ReLU). 
Then, we truncate negative values in the activation map $h^{lk}_x$, and multiply it (element-wise) by the truncated gradient map. This ensures that only pixels associated with both positive activation and gradients are preserved.
Then, we sum the saliency maps across the channel (filter) axis to aggregate $saliency$ per pixel from all channels in the $l$-th layer. The $l$-th saliency map $m^l_x$ is obtained by resizing (via bi-cubic interpolation) to the original image spatial dimensions followed by min-max normalization. This process produces a set of $L$ saliency maps ${M=\{m^l_x\}_{l=1}^L}$.

The final saliency map ${z_x^n \triangleq z(M,n)}$ is computed based on a function $z$ that aggregate the information from the saliency maps produced by last $n$ layers. 
In this work, we implement $z$ as follows 
\begin{equation}
\label{eq:final-heatmap}
  z(M,n)=\frac{1}{n}\textstyle\sum_{l=L-n+1}^L m^l_x.
\end{equation}
Note that in our experiments, we found out that different implementations of $z$, such as max-pooling, Hadamard product, or various weighted combinations of $M$, performs worse than Eq.~\ref{eq:final-heatmap}. Yet, in Sec.~\ref{sec:results}, we do investigate the effect of different $n$ values on the final saliency map $z_x^n$.

\subsection{GAM's Unique Features}
\label{subsec:gam-vs-gc}

\begin{figure}[t]
    \centering
    \includegraphics[width=\columnwidth]{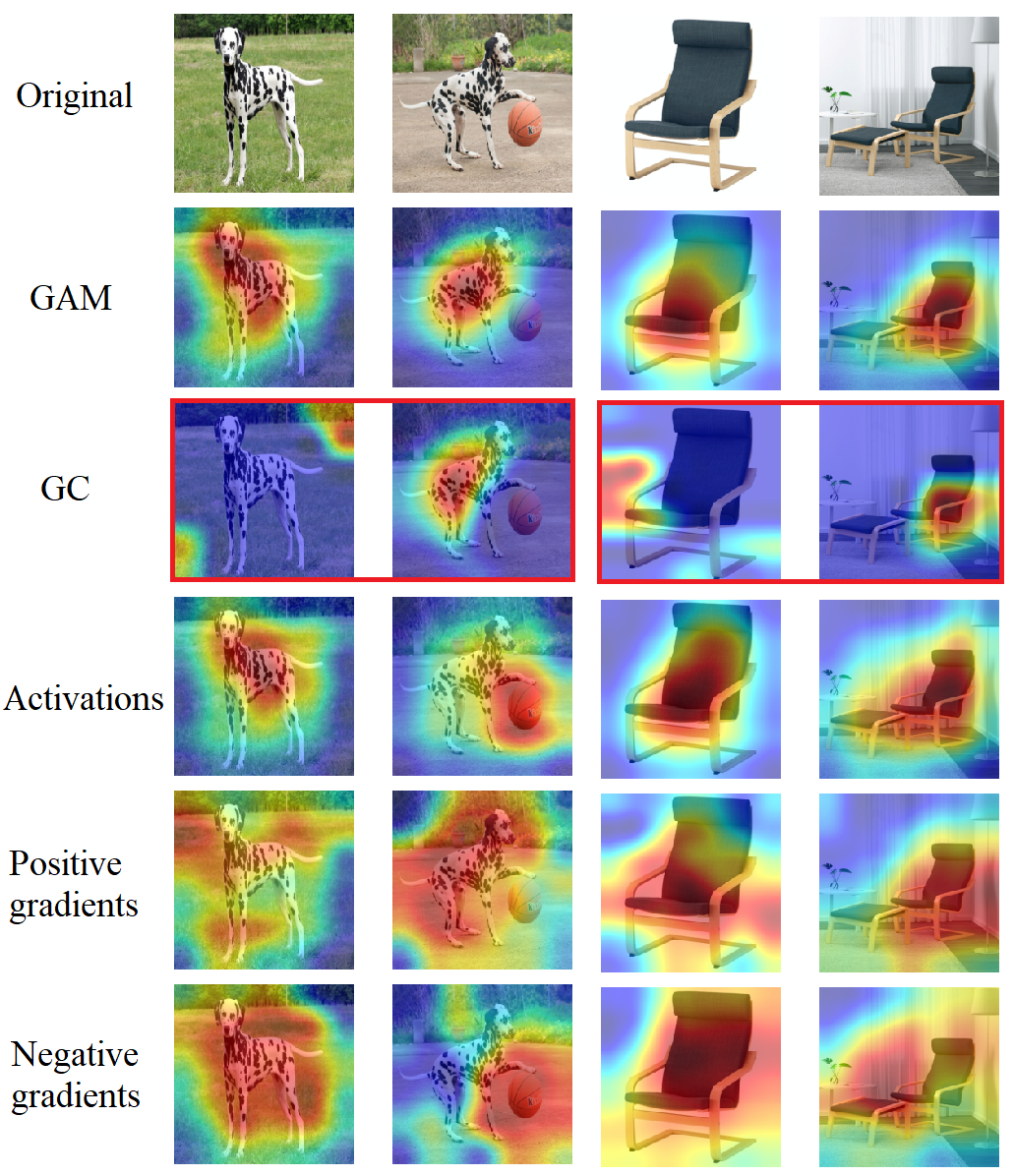}
    \vspace{-4mm}
    \caption{GAM and GC saliency maps w.r.t. the cosine similarity for two pairs of images: \emph{Dogs} and \emph{Chairs} (DenseNet201). GC's failures are marked red. Rows 4-6 present the activation map, ReLUed positive gradient maps (summed across channels), and negative gradient map (summed across channels after negation and ReLU), respectively. See Sec.~\ref{subsec:gam-vs-gc} for details.}     \label{fig:neg_grads}
\end{figure}

GAM presents several advantages over GC and GC++:

\textbf{Gradient Localization:} GC computes the saliency map based on a linear combination of the activation maps in the last convolutional layer as follows:
\begin{equation}
\label{eq:gc-heatmap}
    m_{GC}(h_x^L, g_x^L) = \textrm{NRM} \left[\textrm{RSZ} \left[\phi\left(\sum_{k=1}^{c_l}\alpha_x^kh_x^{Lk}\right)\right]\right],
\end{equation}
where $\alpha_x^k=\frac{1}{u_Lv_L}\sum_{i=1}^{u_L} \sum_{j=1}^{v_L}[g_x^{Lk}]_{ij}$. 
When compared to GAM, the computation in Eq.~\ref{eq:gc-heatmap} has two major drawbacks:
First, the coefficients $\alpha_x^k$ are the \emph{pooled} gradients. Hence, in GC (and GC++), the gradient spatial information is lost. This is in contrast to our GAM approach (Eq.~\ref{eq:l-heatmap}) that preserves (positive) gradient localization via the element-wise multiplication by $\phi(g_x^{lk})$. The significance of this property is well expressed in the \emph{Positive gradients} row in Fig.~\ref{fig:neg_grads}.

\textbf{Multi-layer Analysis:} GC produces saliency maps based on the last convolutional layer only. GAM, on the other hand, gleans information extracted from \emph{multiple} layers (or blocks) that vary by their resolution and sensitivity (Eq. \ref{eq:final-heatmap}). 
Earlier blocks in the network are characterized with higher resolution. For example, in DenseNet, the last convolutional layer produces low-resolution activation maps of size $7 \times 7$ whereas the preceding convolutional layer produces activation maps of  $14 \times 14$.

Our findings show that extracting information from earlier blocks is critical in certain architectures. In Sec.~\ref{sec:results}, we show that incorporating information from earlier blocks (i.e, setting $n>1$) enables GAM to produce fine-grained saliency maps that are more focused on the relevant objects. However, the application of the same feature to GC / GC++ hurts performance (Fig. \ref{fig:multilayer_comp} and Tabs.~\ref{tab:objective}, \ref{tab:loc}).

\textbf{Negative Gradients Suppression:} A subtle, yet highly important drawback of Eq.~\ref{eq:gc-heatmap} stems from the way in which the $\phi$ (ReLU) operation is applied. In GC, the weighted combination of the activations $h_x^{Lk}$ is summed, where each activation is weighted by its pooled gradients $\alpha_x^k$.
In architectures like ResNet or DenseNet, $h_x^{Lk}$ are always non-negative (due to the ReLU activation at the end of each block). However, the pooled gradients can still result in a negative value. 
As a result, GC might become insensitive to important regions (pixels) that should be intensified. The justification for this claim is as follows: Consider a pixel ${(i,j)}$ in a region that contributes to the final score $s$. Ideally, we wish this pixel to be intensified in the final saliency map. By its nature, such a pixel in an ``important'' region is expected to have positive (pooled) gradient values and positive activation values across several filters. However, it is also possible that some other filters that respond with a small, yet positive activation, will be associated with negative (pooled) gradients values. Mathematically, this is expressed by the following decomposition: 
\begin{equation}
    \label{eq:gc_problem}
      \underbrace{[\sum_{k=1}^{c_l}\alpha_x^kh_x^{Lk}]_{ij}}_{A_{ij}} = \underbrace{[\sum_{k:\alpha_x^k<0}\alpha_x^kh_x^{Lk}]_{ij}}_{N_{ij}} + \underbrace{[\sum_{k:\alpha_x^k\geq0}\alpha_x^kh_x^{Lk}]_{ij}}_{P_{ij}}.
\end{equation}
If ${|N_{ij}| \geq P_{ij}}$, then the pixel $(i,j)$ will have an intensity ${A_{ij} \leq 0}$.
In this case, the pixel ${(i,j)}$ as well as other pixels in the region, are zeroed and masked due to the subsequent application of $\phi$ (ReLU) in Eq. \ref{eq:gc-heatmap}. This might further lead to a relative intensification of other, less ``informative'' pixels $(i',j')$ (associated with much smaller contributions than of ${(i,j)}$), but for which ${A_{i'j'}>0}$.
    
GAM on the other hand, applies $\phi$ to the gradients $g^{lk}_x$ \emph{before} the multiplication by the activations $h^{lk}_x$ (Eq.~\ref{eq:l-heatmap}). This ensures negative gradients are zeroed and hence do not (negatively) affect the region's intensity on other channels or layers. Thus, regions with positive gradients are never masked by $\phi$ and ``correctly'' intensified according to the magnitudes of the positive gradients and activations only.

In GC, the negative gradients problem becomes noticeable when using the cosine similarity. Fig.~\ref{fig:neg_grads} exemplifies this effect, presenting a comparison between GC and GAM (using DenseNet201). We used the `last layer' version of GAM (Eq. \ref{eq:final-heatmap}, ${n=1}$), ensuring the improvement by GAM is indeed due to the way it computes the saliency maps, neutralizing the contribution from earlier layers. Each pair of columns in Fig. \ref{fig:neg_grads} presents saliency maps computed w.r.t. the cosine similarity. We see that GC (third row, marked red) produces saliency maps that intensify wrong regions (left image of each pair in rows 2). Empirically, this is explained by the accumulated activation maps and the positive gradient maps (shown in rows 4-5 after ReLU), and the negative gradient maps (shown in row 6 after negation and ReLU). In both examples (dog and chair), we observe the high magnitude of the negative gradients and their adverse effect: the final intensity in regions of interests is significantly attenuated compared to the background, resulting in poor quality saliency maps. However, as explained above, by suppressing negative gradients in advance, GAM averts this problem and successfully produces adequate saliency maps.

\begin{figure}
    \centering
    \includegraphics[width=\columnwidth]{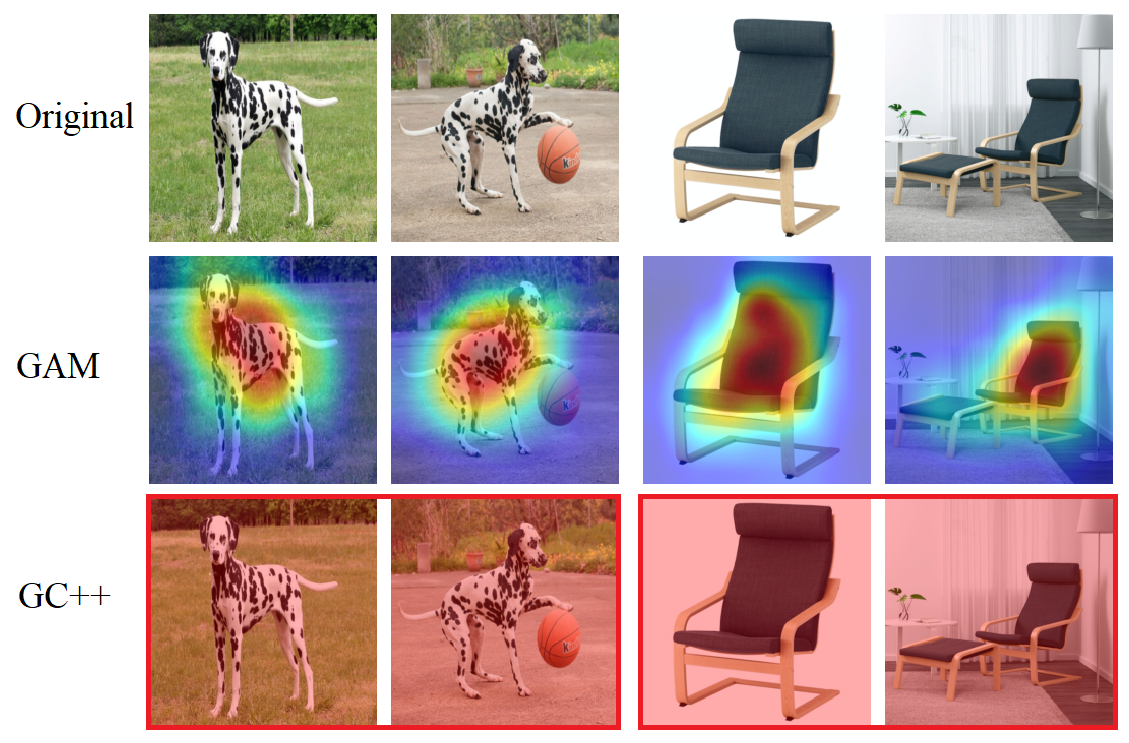}
    \vspace{-3mm}
    \caption{
    GAM and GC++ saliency maps w.r.t. dot product similarity for two pairs of images (DenseNet201). GC++'s failures are marked red. See Sec.~\ref{subsec:gam-vs-gc} for details.
    } 
    \label{fig:gradcampp_fail}
    \vspace{-2mm}

\end{figure}

The poor performance of GC, when using the cosine similarity (instead of the dot-product), can be further explained mathematically: In the case of the dot-product similarity, ${s(f_x,f_y)=f_x^Tf_y}$, and the gradients $g^{lk}_x, g^{lk}_y$ are guaranteed to be non-negative. This stems from the fact that in DenseNet (and many other architectures), the global average pooling operation is applied \emph{after} the application of ReLU, hence both ${\frac{\partial s}{\partial f_x}=f_y}$ and ${\frac{\partial s}{\partial f_y}=f_x}$ are entry-wise non-negative, and so does $g^{lk}_x$ and $g^{lk}_y$ (as the gradient of the average pooling function is a positive constant). This implies ${N_{ij}=0}$ for all $(i,j)$ in Eq. \ref{eq:gc_problem}, thus negative gradients do not exist at all. 
However, in the case of the cosine similarity, ${s(f_x,f_y)=\frac{f_x^Tf_y}{||f_x||||f_y||}}$), and since both $f_x$ and $f_y$ are entry-wise non-negative we have ${s(f_x,f_y)\geq 0}$ and:
\begin{equation}
\label{eq:cos-grad}
    \frac{\partial s}{\partial f_x}=\underbrace{\frac{f_y}{||f_x||||f_y||}}_{ \text{ entry-wise} \geq 0} - \underbrace{s(f_x,f_y)}_{\geq 0}\underbrace{\frac{f_x}{||f_x||^2}}_{\text{ entry-wise} \geq 0}.
\end{equation}
Eq.~\ref{eq:cos-grad} shows that $\frac{\partial s}{\partial f_x}$ (and hence $\frac{\partial s}{\partial f_y}$) is the difference between two positive vectors, and hence may contain negative entries. Therefore, in the case of cosine similarity, negative gradients are possible, and might mask "important" regions in the image that should be intensified in the saliency map.

Finally, when using the dot product similarity, it is GC++ that completely fails. Fig.~\ref{fig:gradcampp_fail} compares between GC++ and the `last layer' GAM (${n=1}$). GC++ weighs the pixel-wise gradients (before pooling) with the coefficients:
\begin{equation}
\label{eq:grad-campp}
    \beta_{ij}^{k}=
    \frac{\frac{\partial^2 \exp(s)}{(\partial h_{ij}^{Lk})^2} }
    {2\frac{\partial^2 \exp(s)}{(\partial h_{ij}^{Lk})^2} +\sum_a\sum_b h_{ab}^{Lk} \frac{\partial^3 \exp(s)}{(\partial h_{ij}^{Lk})^3  }}.
\end{equation}
Note that during the computation of $\beta_{ij}^{k}$, GC++ passes $s$ through the exponential function. However, when $s$ is the dot-product, this may lead to an "explosion" of the saliency map values, as observed in Fig.~\ref{fig:gradcampp_fail}. GAM, however, produces adequate saliency maps.


\begin{figure*}[t]
    \centering
    \includegraphics[width=\textwidth]{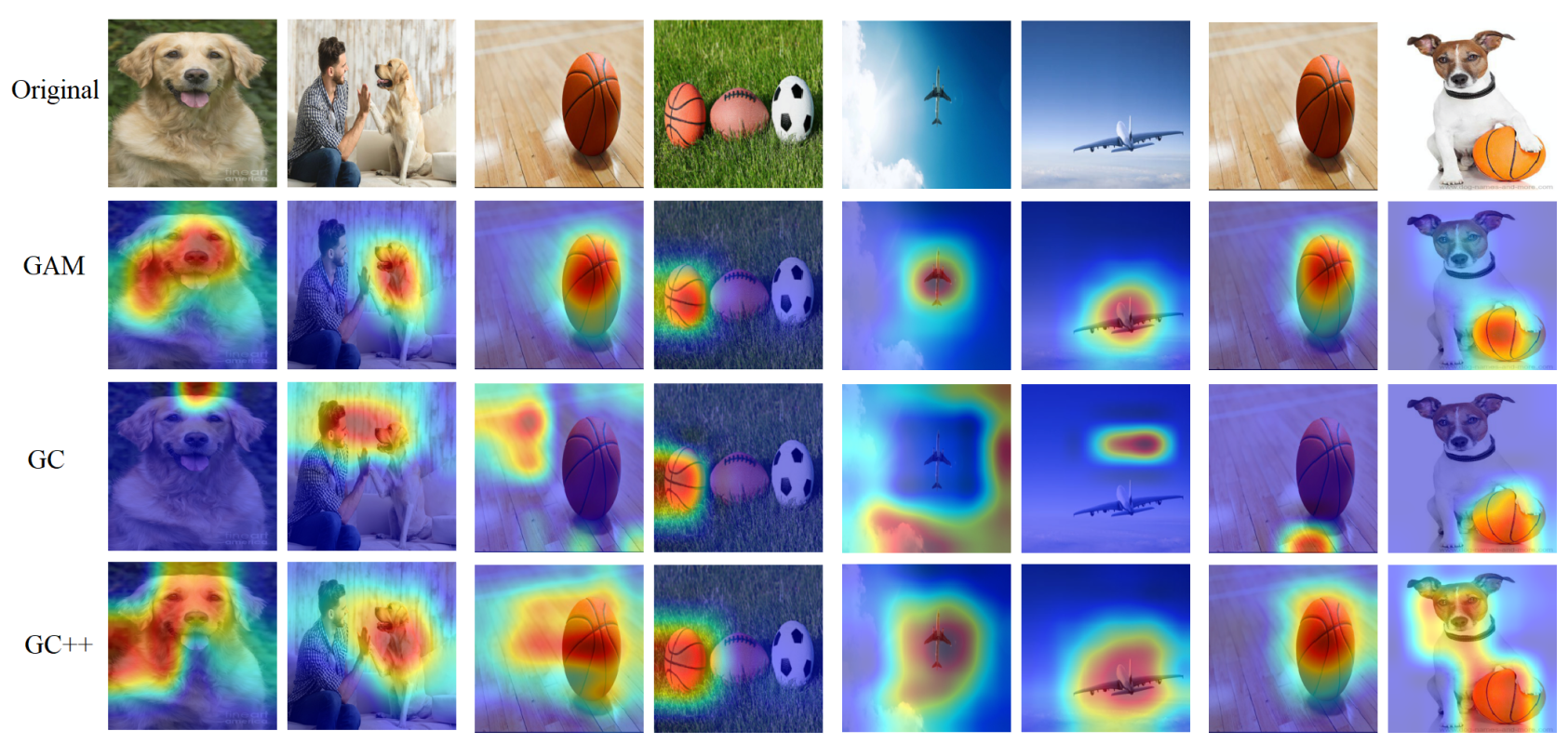}
    \caption{Each pair of rows presents saliency maps produced by GAM, GC and GC++ w.r.t. the cosine similarity.}
    \label{fig:sim_densnet_cos}
\end{figure*}

\begin{figure*}
    \centering
    \begin{overpic}[width=\textwidth]{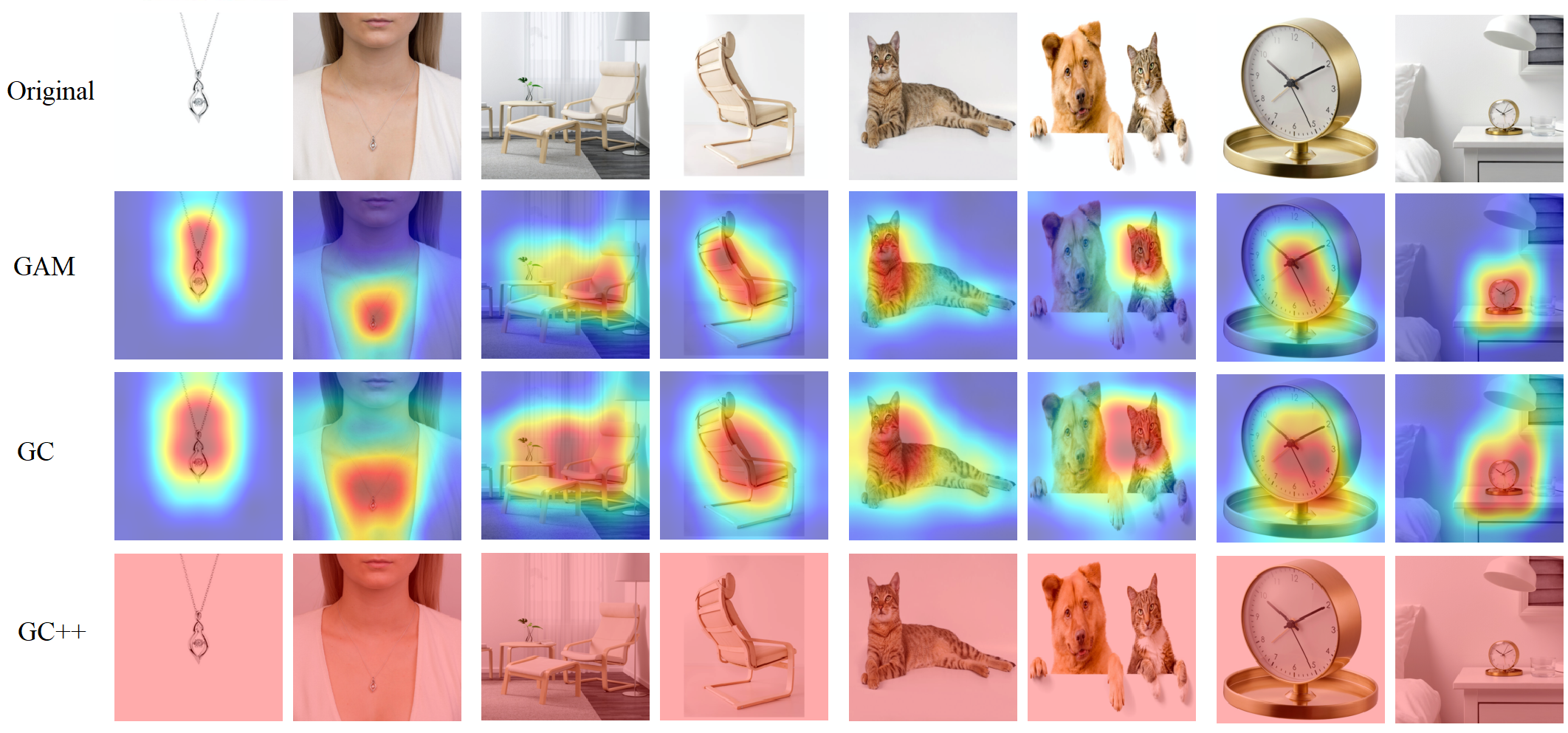}
    \put(30,1){\color{black}\linethickness{0.2mm}\dashline[50]{1.5}(0,0)(0,45)}
    \put(53.42,1){\color{black}\linethickness{0.2mm}\dashline[50]{1.5}(0,0)(0,45)}
    \put(76.86,1){\color{black}\linethickness{0.2mm}\dashline[50]{1.5}(0,0)(0,45)}
    
    \end{overpic}
    \caption{Saliency maps produced by GAM, GC and GC++ (DenseNet201 model) w.r.t. the dot-product similarity. Each pair of columns corresponds to a pair of images for which the similarity score was computed.}
    \label{fig:sim_dot}
\end{figure*}

\section{Experimental Results}
\label{sec:results}

\subsection{Subjective Evaluation}
\label{subsec:res_sim}
First, we demonstrate GAM's ability to explain visual similarity models. To this end, we set $f$ to the embedding produced by the (channel-wise) global average pooling layer in an ImageNet pretrained DenseNet201 model (discarding the classifier head). To determine the similarity of two images, $x$ and $y$, the images are passed through the model to generate the embeddings $f_x$ and $f_y$. Then, the similarity score is computed by $s(f_x,f_y)$, where $s$ is either the dot-product or cosine similarity.

In Fig.~\ref{fig:sim_densnet_cos}, row-pairs present saliency maps for pairs of image representations w.r.t. the cosine similarity. The saliency maps by GAM, were produced using two layers (setting $n=2$ in Eq.~\ref{eq:final-heatmap}). We see that GAM produces quality saliency maps, while GC (column 3) consistently fails. When compared to GC++, GAM exhibits saliency maps that are more focused on the source for the similarity. Results w.r.t. the dot-product appear in Fig.~\ref{fig:sim_dot}. In this case, we see that GC++ completely fails.

\begin{figure}[t]
    \centering
    \includegraphics[width=0.9\columnwidth]{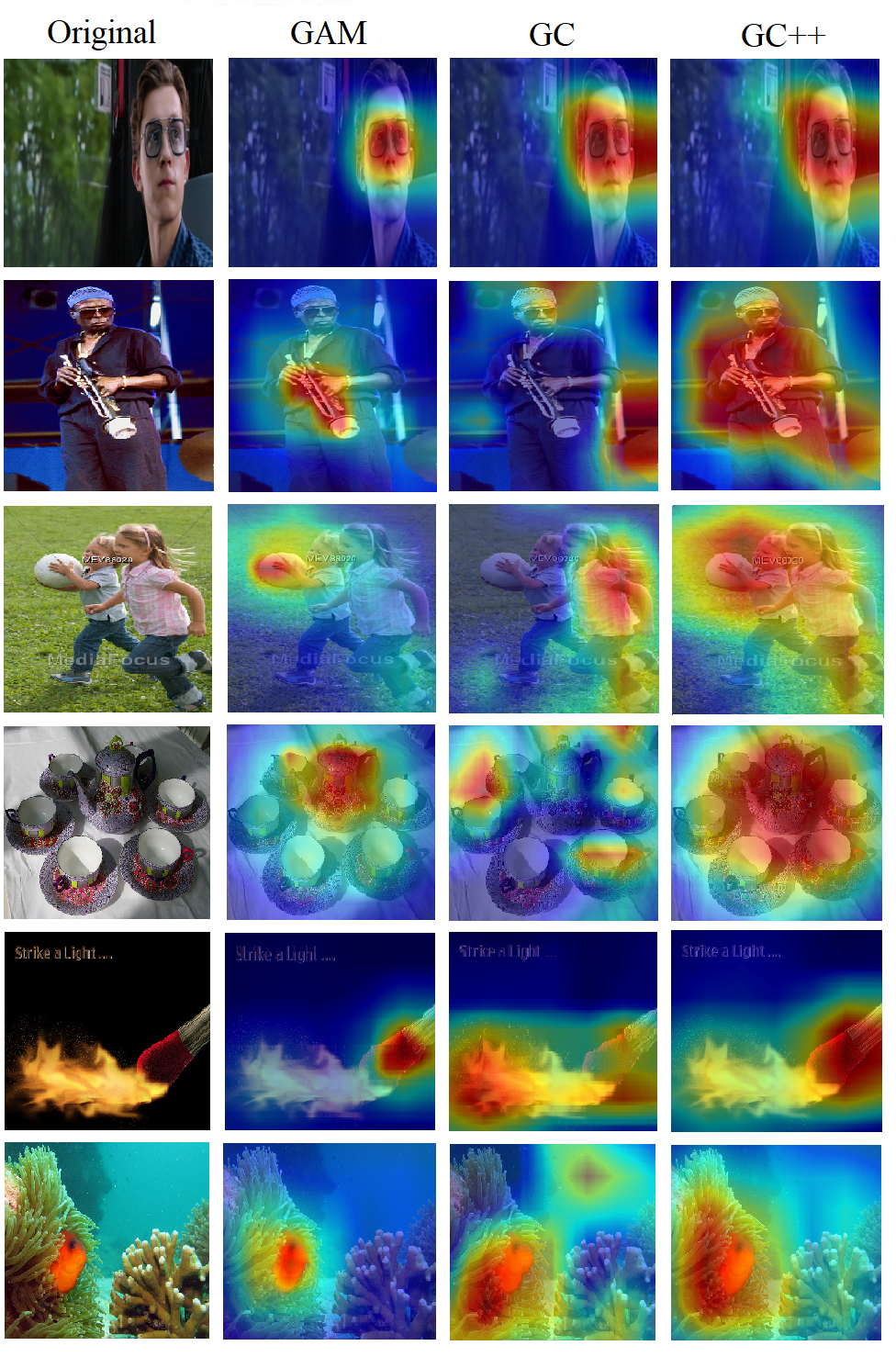}
    \vspace{-5mm}
    \caption{Saliency maps produced by GAM, GC and GC++ w.r.t. the classes (top to bottom) "sunglasses", "oboe", "soccer ball", "coffeepot", "matchstick" and "anemone fish".}
    \label{fig:classification}
    \vspace{-7mm}
\end{figure}

Next, we turn to demonstrate GAM's ability to visually explain classification models. In this case, the saliency maps are computed w.r.t. the logits scores produced by DenseNet-201. Specifically, we compute $s(f_x,w_j)$, where $s$ is the dot-product, $f_x$ is the image representation, and $w_j$ is the weights vector associated with the class $j$. Fig.~\ref{fig:classification} presents examples of saliency maps produced by GAM ($n=2$), GC and GC++. It is visible that GAM produces saliency maps that are more class discriminative than the ones produced by GC and GC++. These results further support the analysis from Sec.~\ref{subsec:gam-vs-gc}, demonstrating the advantages of GAM (over GC and GC++), and show that GAM generates adequate saliency maps in all settings.

\subsection{Sanity Checks for Saliency Maps}
\label{sec:res_sanity}
As explained in Sec.~\ref{subsec:related-sanity}, visually appealing saliency maps can be misleading. To assess the validity of GAM for explanations, we conduct the \emph{parameter randomization} and the \emph{data randomization} sanity tests from \cite{adebayo2018sanity}. GAM passed both tests.

\begin{figure}
    \centering
    \includegraphics[width=\columnwidth]{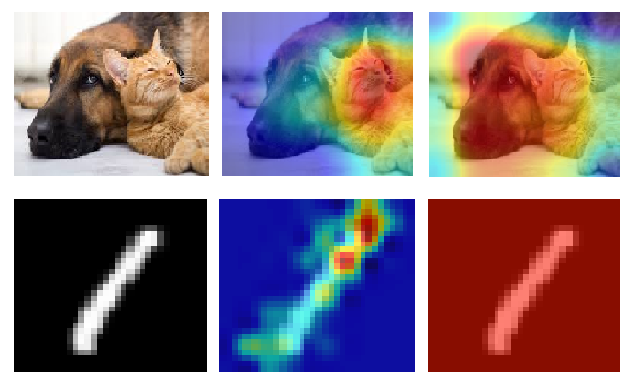} \vspace{-3mm}
    \caption{Sanity checks. Rows 1 and 2 present GAM results for the \emph{parameter randomization} and \emph{data randomization} tests w.r.t. the ``tabby cat'' (ImageNet) and ``one'' (MNIST) classes, using ResNet50 and LeNet-5, respectively. Left to right: Row 1: Original image, GAM computed based on a trained model, GAM computed based on an untrained model (random weights). Row 2: Original image, GAM computed based on a model that was trained with the ground truth labels, GAM computed based on a model that was trained with random labels.}
    \label{fig:sanity_checkes}
\end{figure}

Figure~\ref{fig:sanity_checkes} presents examples from the sanity checks. The first row shows two saliency maps produced by GAM w.r.t. the ``tabby cat'' class. We see that when GAM utilizes an ImageNet pretrained ResNet50 model, it produces a focused saliency (around the cat), but when applying GAM to the same network with randomly initialized weights, it fails to detect the cat in the image. Thus, we conclude that GAM is sensitive to model parameters and passes the \emph{parameter randomization} test. The second row shows that GAM produces an adequate saliency map when the model (LeNet-5~\cite{lecun1998gradient}) is trained with the true MNIST labels, but fails when the model is trained with random labels. Thus, we conclude that GAM is sensitive to data labels and passes the \emph{data randomization} test.


\begin{figure}
    \centering
    \includegraphics[width=\columnwidth]{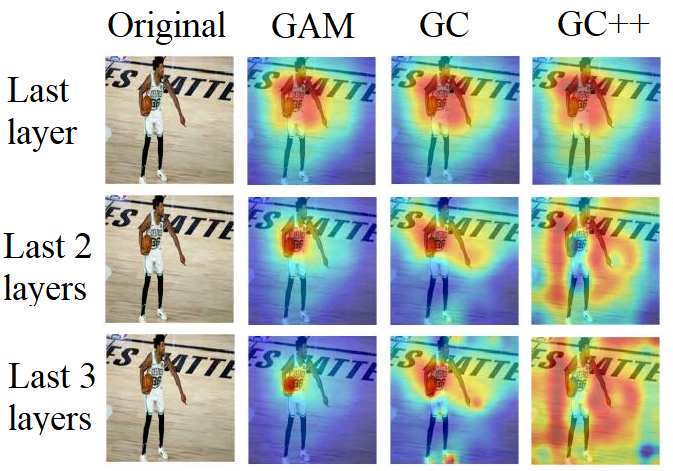}
    \caption{Layer ablation study (DenseNet201). Saliency maps are computed by GAM, GC and GC++, for $n=1,2,3$(Eq. \ref{eq:final-heatmap}), w.r.t. to class "basketball". GAM performs the best. See Sec.~\ref{subsec:ablation}.}
    \label{fig:multilayer_comp}
\end{figure}

\begin{figure*}
    \centering
    \includegraphics[width=.7\textwidth]{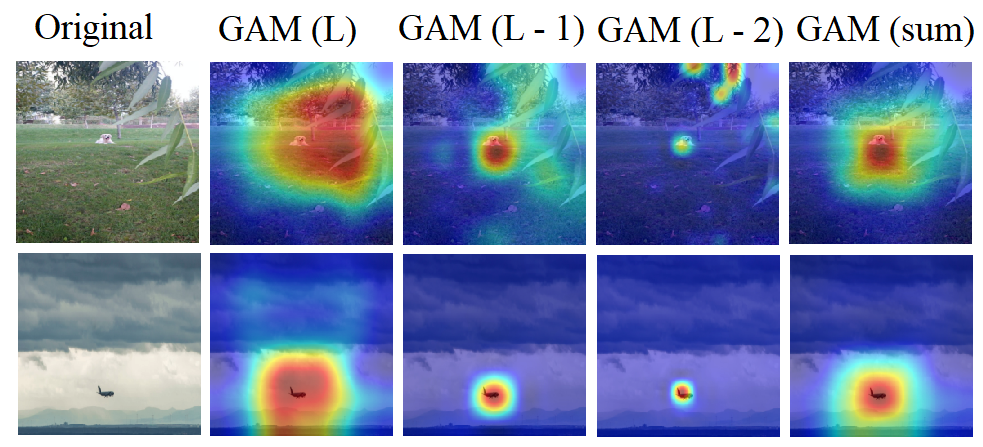}
    \caption{GAM for small objects (DenseNet201). Saliency maps are computed w.r.t. the classes "golden retriever" (row 1) and "airliner" (row 2), for each layer $l=L, L-1, L-2$ and their sum (Eq. \ref{eq:final-heatmap}, $n=3$).}
    \label{fig:small_extra}
\end{figure*}

\begin{figure*}
    \centering
    \includegraphics[width=.9\textwidth]{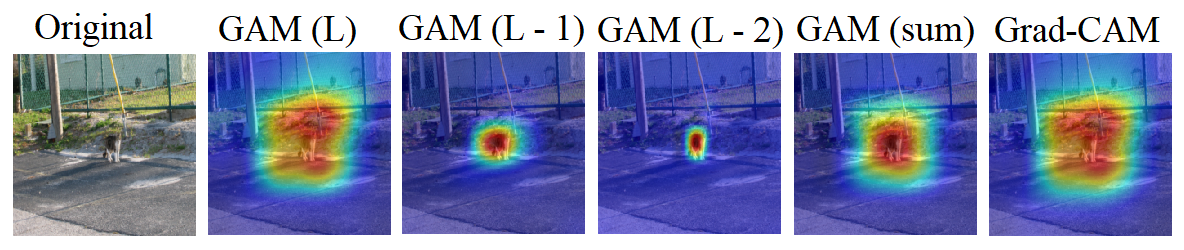}
    \caption{GAM for small objects (DenseNet201). Saliency maps are computed w.r.t. the classes "tabby cat", for each layer $l=L, L-1, L-2$ and their sum (Eq. \ref{eq:final-heatmap}, $n=3$). The last column presents results produced by GC.}
    \label{fig:small1}
\end{figure*}

\subsection{Layer Ablation Study}
\label{subsec:ablation}
In this section, we test whether GAM, GC, and GC++ benefit from the use of multiple layers. On one hand, earlier layers are associated with smaller receptive fields, giving better localization. On the other hand, these layers usually account for less semantic features. 

Fig.~\ref{fig:multilayer_comp} presents a comparison of GAM, GC and GC++ when using multiple layers ($n\in\{1,2,3\}$). We see that GAM benefits from the use of multiple layers, while GC and GC++ do not.

Figures~\ref{fig:small_extra} and \ref{fig:small1} demonstrate the advantage of using multi-layer GAM compared to a single layer GAM. Three images are presented, each with a small object (dog, airplane, and cat). We see that GAM based on earlier layers ($l=L-1, L-2$) produces more focused saliency maps due to higher resolution analysis. This leads to a better localization in the final saliency map as seen in \emph{'GAM(sum)'} ($n=3$). 

\begin{figure*}
    \centering
    \includegraphics[width=\textwidth]{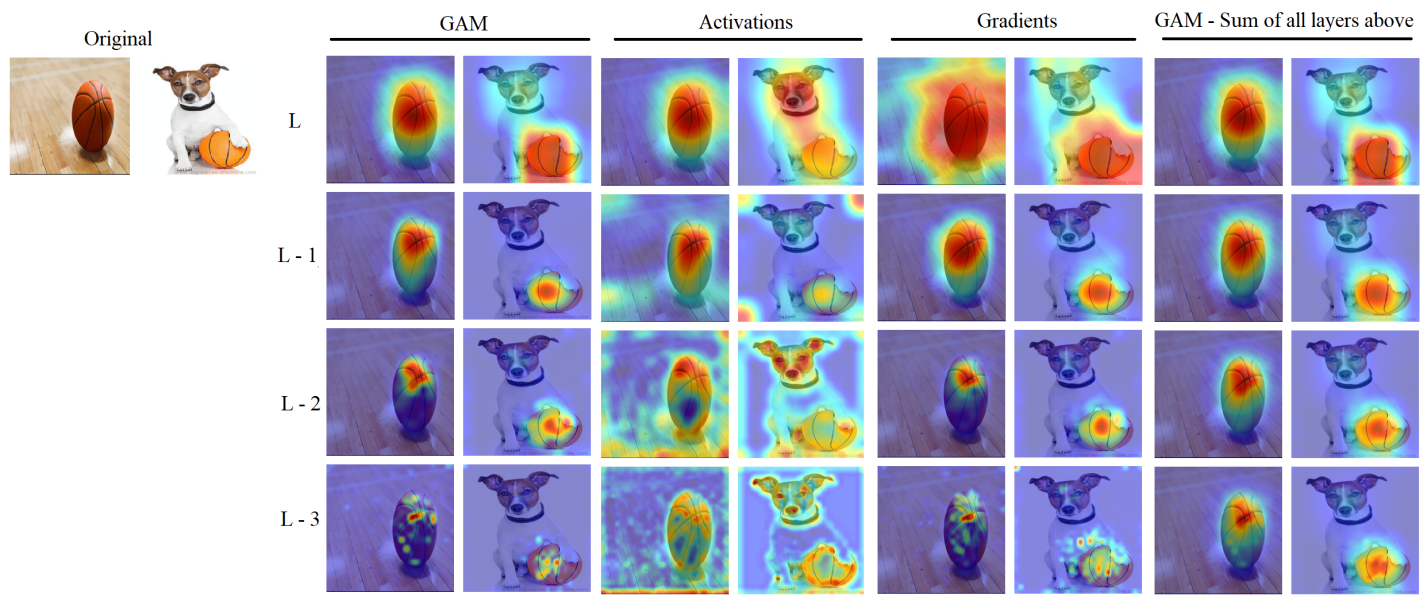}
    \caption{GAM for visual similarity using pretrained Imagenet Densent201:  Layer ablation study. Columns 1-2, 3-4 and 5-6 present the saliency maps $m^l_x$ (Eq.~1), activation maps $h^l_x$ (summed over the channel axis) and gradient maps $\phi(g^l_x)$ (summed over channels), respectively, for $l=L,L-1,L-2, L-3$ (top to bottom). The last two columns present saliency maps computed based on, Eq.~2, with $n=1,2,3,4$ (top to bottom), respectively.}
    \label{fig:dens_layers}
\end{figure*}

Figure~\ref{fig:dens_layers} presents another layer-wise analysis, where it is observed that the last two layers (second row, last two columns), corresponding to $n=2$, best balances localization with the extraction of semantic features, yielding optimal results. In addition, gradient localization is observed in the 'Gradients' columns, which is a unique property of GAM (in contrast to GC that performs gradient pooling). For further explanations, see Sec.~\ref{subsec:gam-vs-gc} (Gradient Localization).

Indeed, in our experiments, we noticed that GAM with $n=2$ best balances localization with the extraction of semantic features. Yet, when setting $n>1$ for GC and GC++, performance degrades. As we shall see, these trends repeat in the quantitative evaluation in Secs.~\ref{subsec:recognition} and \ref{subsec:loc} as well.

\begin{table*}
\centering
\large
\vskip 0.1in
\begin{tabular}{l@{\hskip -.2cm}c|ccc|ccc|ccc}
\toprule
 \textbf{Task} & \textbf{Metric} & &\textbf{GAM} & & & \textbf{GC++} & & & \textbf{GC} & \\
 & & 1 & 2 & Impr.& 1 & 2 & Impr.& 1 & 2 & Impr.  \\
\midrule
\textbf{Classification} & & & & & & & & & & \\
VRC & ADP ($\downarrow$) &
17.47 & \textbf{17.22} & 1.4\% &
17.62 & 17.67 & -0.3\% &
18.49 & 18.56 & -0.4\%  \\
VRC (25\%) & ADP ($\downarrow$) &
18.57 & \textbf{18.51} & 0.3\% &
19.31 & 19.45 & -0.3\% &
20.32 & 20.37 & -0.2\% \\
VRC (10\%) & ADP ($\downarrow$) &
21.02 & \textbf{20.52} & 2.4\% &
22.39 & 22.54 & -0.7\% &
24.89 & 25.23 &  -1.3\% \\
VRC & PIC ($\uparrow$) &
38.12 & \textbf{39.53} & 3.6\%&
37.99 & 35.76 & -6.2\% &
35.24 & 33.45 & -5.6\%  \\
VRC (25\%) & PIC ($\uparrow$) &
36.87 & \textbf{37.56} & 1.8\% &
35.32 & 35.12 & -0.6\% &
34.70 & 34.54 &  -0.5\% \\
VRC (10\%) & PIC ($\uparrow$) &
35.21 & \textbf{35.48} & 0.8\% &
32.75 & 31.98 & -2.4\% &
32.01 & 31.03 & -3.2\%  \\
\midrule
\textbf{Similarity (cos)} & & & & & & & & & & \\
VRC & ADP ($\downarrow$) &
0.75 & \textbf{0.72} & 2.8\% &
0.75 & 0.79 & -5.1\% &
3.21 & 3.46 & -7.2\%  \\
VRC (25\%)& ADP ($\downarrow$) &
1.10 & \textbf{1.03} & 6.8\% &
1.12 & 1.14 & -1.6\% &
9.65 & 10.67 & -9.6\% \\
VRC (10\%)& ADP ($\downarrow$) &
1.39 & \textbf{1.31} & 6.1\% &
1.51 & 1.67 & -9.6\% &
12.19 & 13.45 &  -9.4\% \\
VRC & PIC ($\uparrow$) &
74.13 & \textbf{75.85} & 2.3\% &
71.76 & 70.43 & -1.9\% &
44.33 & 42.12 &  -5.2\% \\
VRC (25\%)& PIC ($\uparrow$) &
64.23 & \textbf{65.44} & 1.8\% &
61.62 & 60.08 & -2.6\%&
39.14 & 39.02 &  -0.3\% \\
VRC (10\%)& PIC ($\uparrow$) &
54.67 & \textbf{55.96} & 2.3\% &
50.83 & 48.87 & -4.0\% &
27.93 & 26.89 &  -3.9\% \\
\midrule
\textbf{Similarity (dot)} & & & & & & & & & & \\
VRC & ADP ($\downarrow$) &
2.15 & \textbf{2.04} & 5.4\% &
53.45 & 55.65 & -4.0\% &
2.16 & 2.39 &  -9.6\% \\
VRC (25\%)& ADP ($\downarrow$) &
2.13 & \textbf{2.04} & 4.4\% &
57.24 & 60.78 & -5.8\% &
2.23 & 2.35 &  -5.1\% \\
VRC (10\%)& ADP ($\downarrow$) &
2.17 & \textbf{2.08} & 4.3\% &
58.02 & 61.23 & -5.2\% &
2.42 & 2.67 &  -9.4\% \\
VRC & PIC ($\uparrow$) &
71.76 & \textbf{72.96} & 1.6\% &
0.21 & 0.20 & -5.0\% &
68.87 & 68.02 & -1.3\%  \\
VRC (25\%)& PIC ($\uparrow$) &
68.97 & \textbf{70.28} & 1.8\% &
0.07 & 0.07 & 0.0\% &
66.12 & 65.23 & -1.4\% \\
VRC (10\%)& PIC ($\uparrow$) &
68.01 & \textbf{68.99} & 1.4\% &
0.02 & 0.02 & 0.0\% &
63.15 & 62.11 &  -1.7\% \\
\bottomrule
\end{tabular}
\caption{Objective evaluation, including Layer ablation study by using $n=1,2$ (Eq.~\ref{eq:final-heatmap}) last layers of ResNet101. For ADP (PIC), lower (higher) is better. VRC stands for ILSVRC-15-val. 25\% and 10\% symbol the subsets of VRC that contain the small objects as explained in Sec.~\ref{subsec:recognition}.
}
\label{tab:objective}
\end{table*}

\begin{table*}

\centering

\large
\vskip 0.1in
\begin{tabular}{l@{\hskip -.5cm}c|ccc|ccc|ccc}
\toprule
  \textbf{Task and} & \textbf{Model} & &\textbf{GAM} & & & \textbf{GC++} & & & \textbf{GC} & \\
  \textbf{Dataset} & & 1 & 2 & Impr.& 1 & 2 & Impr.& 1 & 2 & Impr.  \\
\midrule
\textbf{Classification} & & & & & & & & & & \\
VRC & DenseNet &
54.9 & \textbf{56.9} & 3.6\%
& 54.9 & 47.7 & -13.1\%
&52.4 & 50.3 & -4.0\%  \\
VRC(25\%) & DenseNet &
39 & \textbf{43.8} & 12.3\% &
39.6 & 20.8 & -47.5\% &
33.5 & 26 & -22.4\%  \\
 VRC(10\%) & DenseNet &
23.4 & \textbf{33} & 41.0\% &
22.6 & 11.5 & -49.1\% &
21.3 & 17.4 & -18.3\%  \\

VRC & ResNet &
55.9 & \textbf{57.1} & 2.1\% &
55 & 53.8 & -4.1\% &
47.8 & 47.2 & -1.3\%  \\
VRC(25\%) & ResNet &
40.8 & \textbf{43.1} & 8.3\% &
40.6 & 38.9 & -4.2\% &
33.6 & 33.5 & -0.3\% \\
VRC(10\%) & ResNet &
26.1 & \textbf{33.4} & 29.5\% &
26.2 & 23.9 & -8.4\% &
23.2 & 22.7 & -2.2\%  \\

XRAY & CheXNet &
25.8 & \textbf{28.4} & 10.1\% &
26.2 & 20.2 & -22.9\% &
24.9 & 21.6 & -13.3\%  \\
\midrule
\textbf{Similarity (cos)} & & & & & & & & & & \\
VRC & DenseNet &
57.4  & \textbf{60.7} & 5.7\% &
57.1 & 52.3  & -10.0\% &
52.8 & 53.5 & 1.3\%  \\
VRC(25\%)& DenseNet &
38.2 & \textbf{41.9} & 9.7\% &
37.4 & 21.9  & -44.4\% &
25.5 & 22.6 & -11.4\%  \\
VRC(10\%)& DenseNet &
31.2 & \textbf{35.7} & 14.4\% &
29.7 & 15.4  & -51.4\% &
18.4 & 16 & -13.0\%  \\
VRC & ResNet &
57.1 & \textbf{58.6} & 2.6\% &
56 & 49.8 & -11.1\% &
39.1 & 38.2 & -2.3\%  \\
VRC(25\%)& ResNet &
38.3 & \textbf{39.3} & 2.6\% &
36.1 & 28.5 & -21.1\% &
27.2 & 24.9 & -8.5\%  \\
VRC(10\%)& ResNet &
31.3 & \textbf{34.9} & 11.5\% &
29.4 & 22.2 & -24.5\% &
21.3 & 20.4 & -4.2\%  \\
\midrule
\textbf{Similarity (dot)} & & & & & & & & & & \\
VRC & DenseNet &
59.2  & \textbf{62.4} & 5.4\% &
1 $>$ & 1 $>$  & -- &
58.9 & 54 & -9.8\%  \\
VRC(25\%)& DenseNet &
39.6 & \textbf{43.8} & 10.6\% &
1 $>$ & 1 $>$  & -- &
38.2 & 25.5 & -36.6\%  \\
VRC(10\%)& DenseNet &
32 & \textbf{36.9} & 15.3\% &
1 $>$ & 1 $>$  & -- &
31.2 & 19.6 & -38.6\%  \\

VRC & ResNet &
57.9 & \textbf{61.9} & 6.9\% &
1 $>$ & 1 $>$  & -- &
57.3 & 57.3 & 0\%  \\
VRC(25\%)& ResNet &
39.3 & \textbf{43.1} & 9.7\% &
1 $>$ & 1 $>$  & -- &
38.6 & 38.1 & -1.3\%  \\
VRC(10\%)& ResNet &
31.9 & \textbf{36.6} & 14.7\% &
1 $>$ & 1 $>$  & -- &
31.2 & 30.5 & -2.3\%  \\
\midrule
\textbf{Segmentation} & & & & & & & & & & \\
COCO & TResNet &
28.3 & \textbf{30.7} & 8.5\% &
27.8 & 27.3  & -0.8\% &
27.2 & 27.5 & 1.1\%  \\
COCO(25\%)& TResNet &
22.7 & \textbf{25.8} & 13.7\% &
21.4 & 21.2  & -0.9\% &
21.5 & 21.1 & -1.9\%  \\
COCO(10\%)& TResNet &
21.4 & \textbf{24.9} & 16.4\% &
20.7 & 20.5 & -1.0\% &
21.1 & 20.3 & -3.8\%  \\
VOC & TResNet &
36.2 & \textbf{38.7} & 6.9\% &
35.5 & 34.8  & -2\% &
35.5 & 34.2 & -3.7\%  \\
VOC(25\%)& TResNet &
34.1 & \textbf{37.2}  & 9.1\% &
32.1 & 31.5  &-1.9\% &
33.5 & 31.1 & -7.2\%  \\
VOC(10\%)& TResNet &
27.1 & \textbf{32.7} & 20.7\% &
26.8 & 25.3 & -5.6\% &
26.2 & 24.9 & -5\%  \\
\bottomrule
\end{tabular}
\caption{Object Localization and segmentation results for different combination of task, dataset, model, and method. For each method, we report the accuracy (IoU\%) achieved by using $n=1,2$ (Eq.~\ref{eq:final-heatmap}) last layers. VRC, XRAY, COCO, and VOC stands for ILSVRC-15, ChestX-ray8, MS-COCO, and Pascal-VOC, respectively. See Sec.~\ref{subsec:loc} for details.}
\label{tab:loc}
\end{table*}

\subsection{Objective Evaluation}

\label{subsec:recognition}

Next, we present objective evaluation, following the measures suggested by in~\cite{chattopadhay2018grad} (we refer to \cite{chattopadhay2018grad} for the full details):

\textbf{\emph{Average Drop Percentage} (ADP)}:
ADP is computed as:
\[\text{ADP}=\frac{100}{N} \sum_{i=1}^N \frac{\max(0,Y_i^c-O_i^c)}{Y_i^c},\]
where $N$ is total number of images in evaluated dataset, $Y_i^c$ is the model's output score (confidence) for the correct class $c$ w.r.t. the original image $i$. $O_i^c$ is the same model’s score, this time w.r.t. the 'explanation map' - a masked version of the original image (produced by the Hadamard product of the original image with the saliency map). The \textbf{lower} the ADP the better the result. 

\textbf{\emph{Percentage of Increase in Confidence} (PIC)}: PIC is computed as:  \[\text{PIC}=\frac{100}{N} \sum_{i=1}^N \mathbbm{1}(Y_i^c<O_i^c).\]
PIC reports the percentage of the cases in which the model’s output scores increase as a result of the replacement of the original image with the explanation map. The \textbf{higher} the PIC the better the result.

We further extended the evaluation from \cite{chattopadhay2018grad} to similarity tasks, by reporting ADP and PIC w.r.t. image-pairs similarity scores (instead of class specific scores). To this end, we created a \emph{similarity subset} (will be made public) by randomly sample image-pairs from the ILSVRC-15-val dataset \cite{ILSVRC15} (which does not overlap with the training set used the trained the models), but with the restriction that each pair contains images that are labeled with the same ground truth class. The \textit{similarity subset} contains $3000$ pairs in total, $3$ for each class.

In addition, we tested the ability of \ourmethod{} to benefit from using several layers, when it is applied on images with small objects. We compare the localization capability on small objects by narrowing the ILSVRC-15-val dataset to a subset that contains images for which the ground truth box area is below the 25\% / 10\% percentile area. For the similarity experiment, we randomly sampled another $3000$ pairs, from the  25\% / 10\%  narrower sets.

The results are reported in Tab.~\ref{tab:objective} (ResNet101). For each method, we report the results both for $n=1$ and $n=2$ (note that $n=3$ performs on par with $n=2$, hence omitted). Recall that for ADP (PIC), lower (higher) values indicate better performance, and Impr. reports the relative improvement obtained by using $n=2$ (over $n=1$). We see that GAM outperforms GC and GC++ at the majority of the scenarios. Moreover, GC (GC++) completely fail when using the cosine (dot-product) similarity. This is another empirical evidence for GC and GC++ limitations (Sec.~\ref{subsec:gam-vs-gc}), and the fact that GAM benefits from multiple layers, whereas GC and GC++ do not (and even degrade). Finally, the results for ResNet101 exhibit the same trends, but are excluded due to space limitation.

\begin{figure}[ht]
    \vspace{10pt}
    \begin{overpic}[width=\columnwidth]{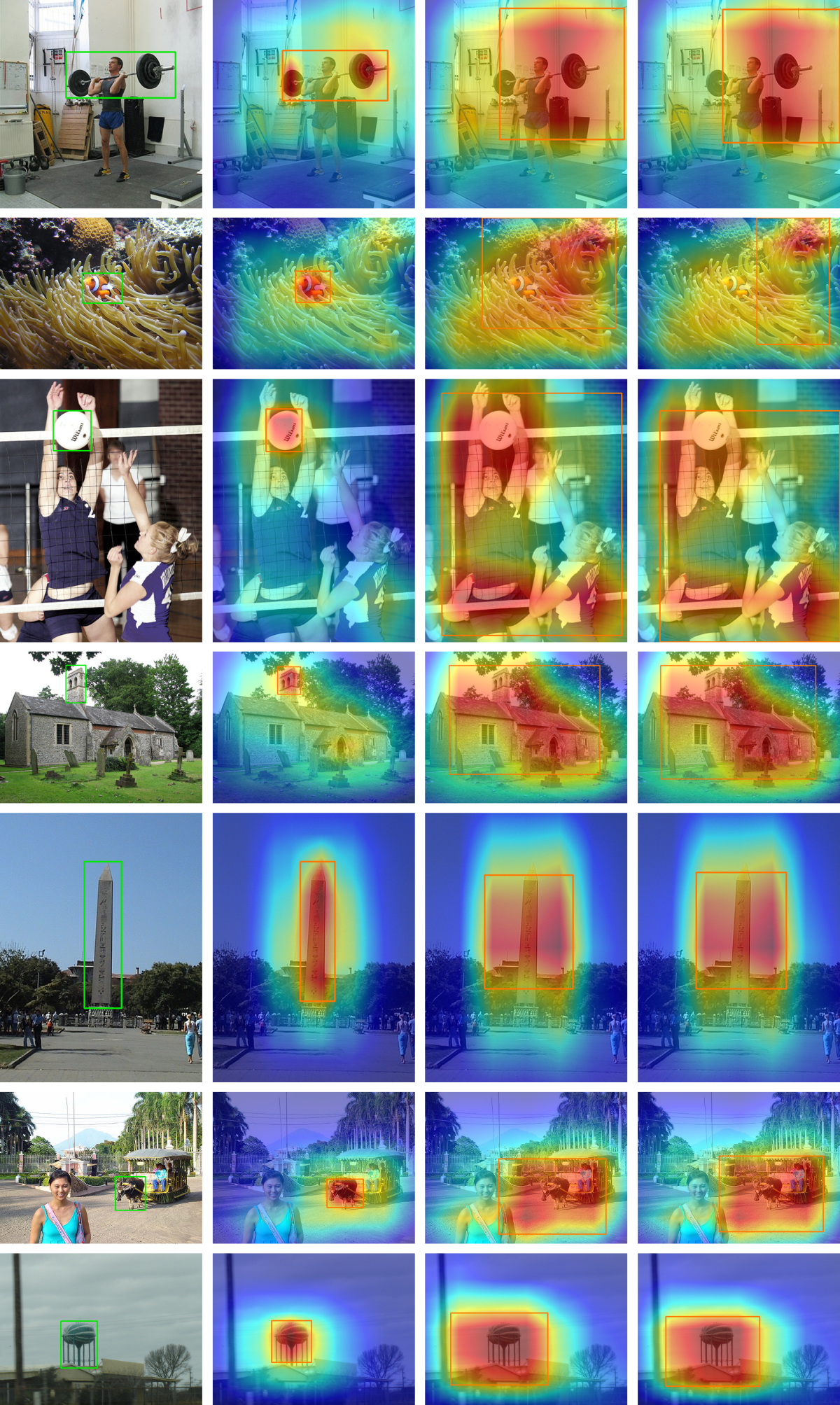}
    \put(2.5, 101){\scriptsize{Ground-truth}}
    \put(20.5,101){\scriptsize{\ourmethod{}}}
    \put(35.5,101){\scriptsize{\gcp{}}}
     \put(50.5,101){\scriptsize{\gc{}}}
    \end{overpic}
    \caption{Object localization via saliency maps using DenseNet201 over ILSVRC-15 dataset, w.r.t. labels: \emph{barbell}, \emph{anemone fish}, \emph{volleyball}, \emph{bell cote}, \emph{obelisk}, \emph{ox} and \emph{water tower}.}
    \label{fig:loc_cls}
    \vspace{-4mm}
\end{figure}

\begin{figure}
    \vspace{10pt}
    \begin{overpic}[width=\columnwidth]{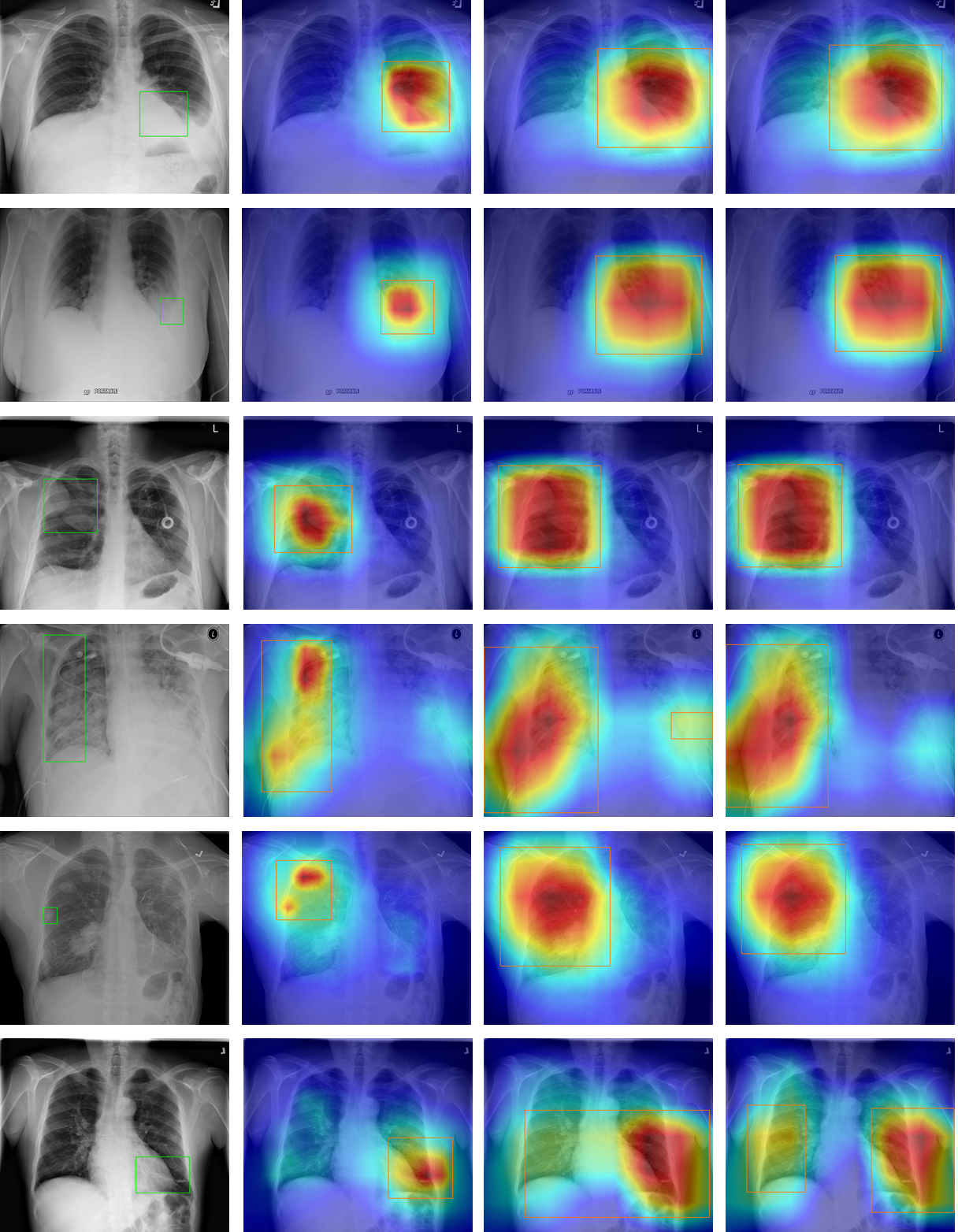}
    
    \put(3.5, 101){\scriptsize{Ground-truth}}
    \put(26,101){\scriptsize{\ourmethod{}}}
    \put(46,101){\scriptsize{\gcp{}}}
     \put(66.5,101){\scriptsize{\gc{}}}
    \end{overpic}
    \caption{Object localization via saliency maps using CheXNet over ChestX-ray8 dataset, w.r.t. pathology: \emph{Atelectasis}, \emph{Effusion}, \emph{Mass}, \emph{Pneumothorax}, \emph{Nodule} and \emph{Pneumonia}. GAM yields saliency maps that are more accurate, hence leading to better localization.}
    \label{fig:loc_chest}
    \vspace{-5mm}
\end{figure}

\begin{figure}
    \begin{overpic}[width=\columnwidth]{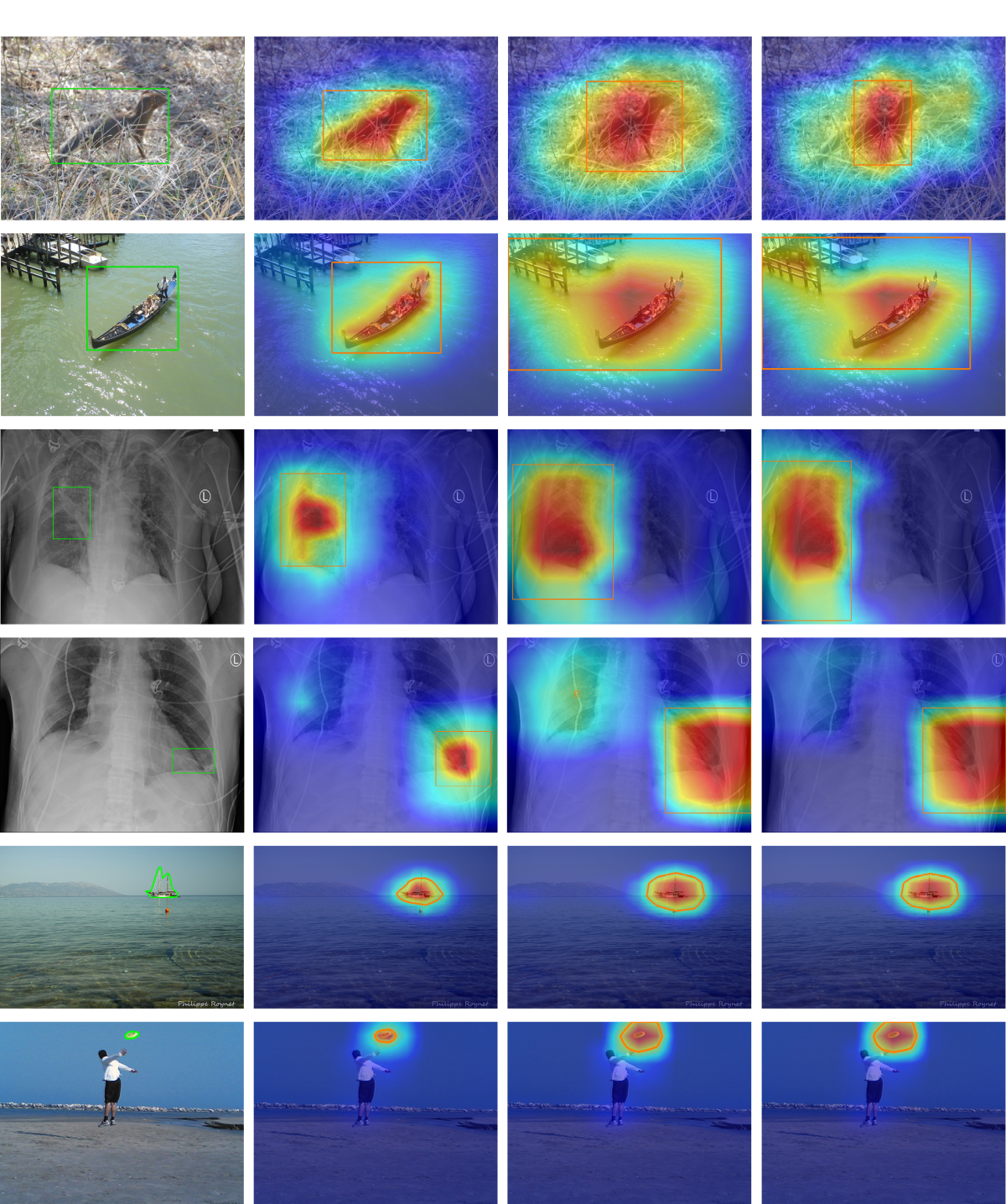}
    \put(4, 98){\scriptsize{Ground-truth}}
    \put(28.5,98){\scriptsize{\ourmethod{}}}
    \put(50,98){\scriptsize{\gcp{}}}
     \put(71.5,98){\scriptsize{\gc{}}}
    \end{overpic}
    \caption{Object localization and segmentation via saliency maps. Rows 1-2, 3-4, 5-6 present BBox generation (orange) using DenseNet201 (w.r.t. labels: \emph{mongoose, gondola}), CheXNet (w.r.t. label: \emph{Atelectasis}), and segmentation (orange) using TResNet (w.r.t. labels: \emph{boat, frisbee}), respectively.}
    \label{fig:loc}
    \vspace{-5mm}
\end{figure}

\subsection{Object Localization and Segmentation}
\label{subsec:loc}
In this section, we compare the localization capability of \ourmethod{}, GC and GC++ via an extensive set of experiments across various tasks, datasets, models, and settings. We measure the quality of the produced saliency maps by Intersection over Union (IoU\%) w.r.t. the ground truth bounding boxes (BBox) or segmented areas. To this end, each saliency map is binarized with a fixed threshold before drawing the predicted BBox or segmented area. The fixed threshold was chosen for each test and method separately by a hold-out set (will be made public).
Table~\ref{tab:loc} presents the obtained localization accuracy (IoU\%) for each combination of task, dataset, model, and method, both for $n=1$ and $n=2$, including the obtained improvement when using $n=2$. Again, we observe that GAM outperforms the other methods. Moreover, it is evident that GAM significantly benefit from using multiple layers (especially in the case of small objects), whereas GC and GC++ suffer from a significant degradation in accuracy when utilizing more than a single layer. In what follows we discuss the results per task.

\textbf{Localization by Classification:}
\label{subsec:detection}
We followed the test protocol from GC~\cite{selvaraju2017grad}, where the saliency maps of a classification model are used to draw a BBox around classified objects.
We apply the two-layer \ourmethod{} (Eq.~\ref{eq:final-heatmap}, $n=2$) \gc{} and \gcp{} on top of pretrained 
DenseNet201 and ResNet101. Figures~\ref{fig:loc_cls}, \ref{fig:loc_chest} and Fig.~\ref{fig:loc} (Rows 1-2) present examples for the generated saliency maps and BBoxes (marked orange). Tab.~\ref{tab:loc} (row 1) presents the localization accuracy (IoU\%) between the predicted and ground truth (ILSVRC-15-val) boxes. In all cases, GAM outperforms both GC and GC++.

\begin{figure}
    \vspace{10pt}
    \begin{overpic}[width=\columnwidth]{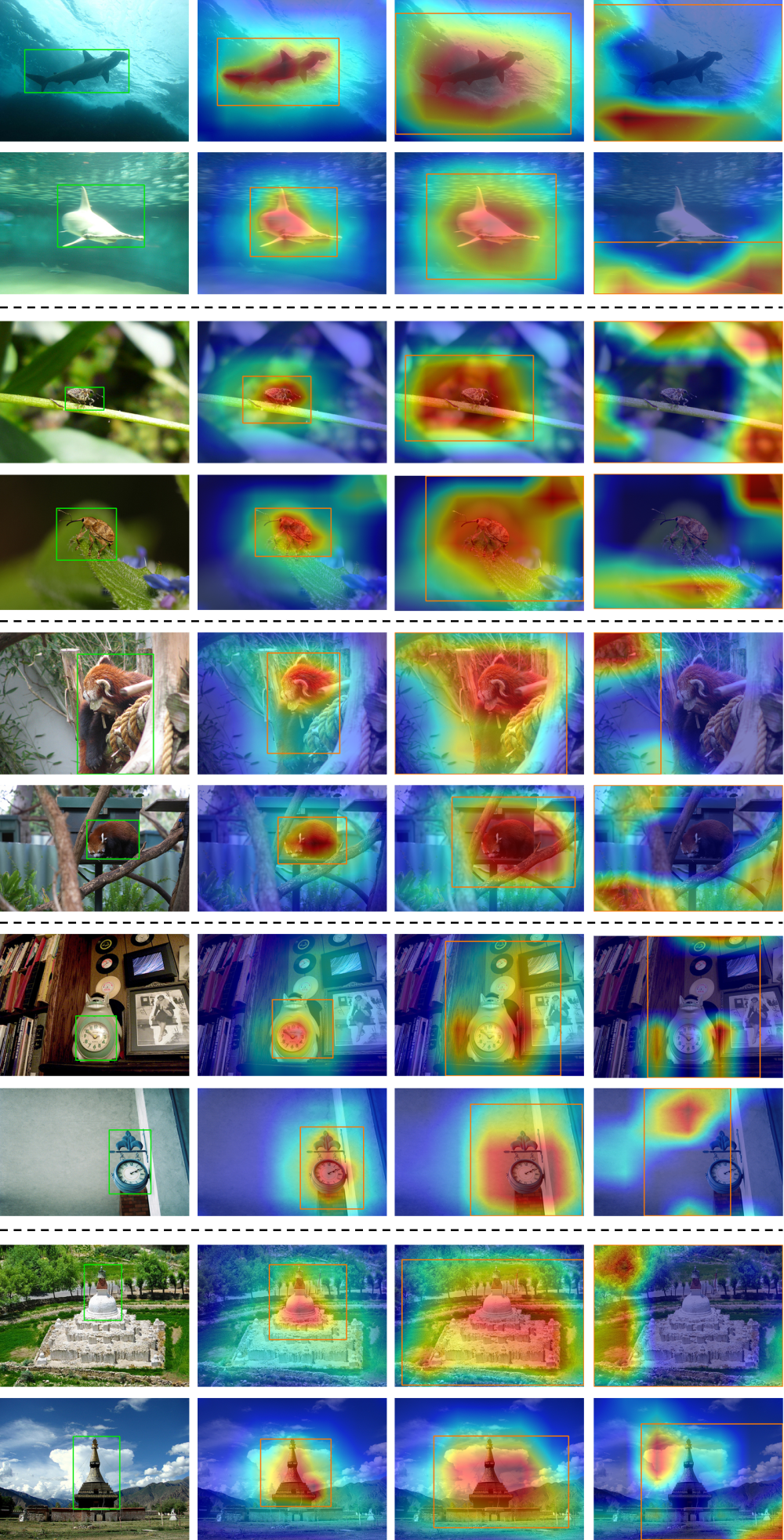}
    \put(2, 101){\scriptsize{Ground-truth}}
    \put(17.5,101){\scriptsize{\ourmethod{}}}
    \put(30,101){\scriptsize{\gcp{}}}
     \put(43,101){\scriptsize{\gc{}}}
    \end{overpic}
    \vspace{-2mm}
    \caption{Object localization w.r.t. similarity score (cosine). The saliency maps are drawn using DenseNet201 over image-pairs from ILSVRC-15-val (validation set). The labels for the image-pairs are (top to bottom): \emph{hammerhead shark}, \emph{weevil}, \emph{lesser panda}, \emph{analog clock} and \emph{stupa}.}
    \label{fig:loc_sim}
    \vspace{-4mm}
\end{figure}

\textbf{Localization by Similarity:}
\label{subsec:detection_sim}
We adjusted the protocol from the localization by classification experiment to support localization by similarity. To this end, we replace the classification score with the similarity score computed for image-pairs. We used the same image-pairs from the \emph{similarity subset} (Sec. \ref{subsec:recognition}). Then, we drew a BBox for each image in the pair, and computed IoU\% w.r.t. to the ground truth. Results w.r.t. the different similarity scores are reported in Tab.~\ref{tab:loc} (rows 5, 8) and demonstrated in Fig.~\ref{fig:loc_sim}. Again, we observe that GC (GC++) fails when using the cosine (dot-product) similarity, and significantly degrades when utilizing multiple layer, while GAM performs the best and clearly benefits from multiresolution analysis.

\begin{figure}
    \vspace{10pt}
    \begin{overpic}[width=\columnwidth]{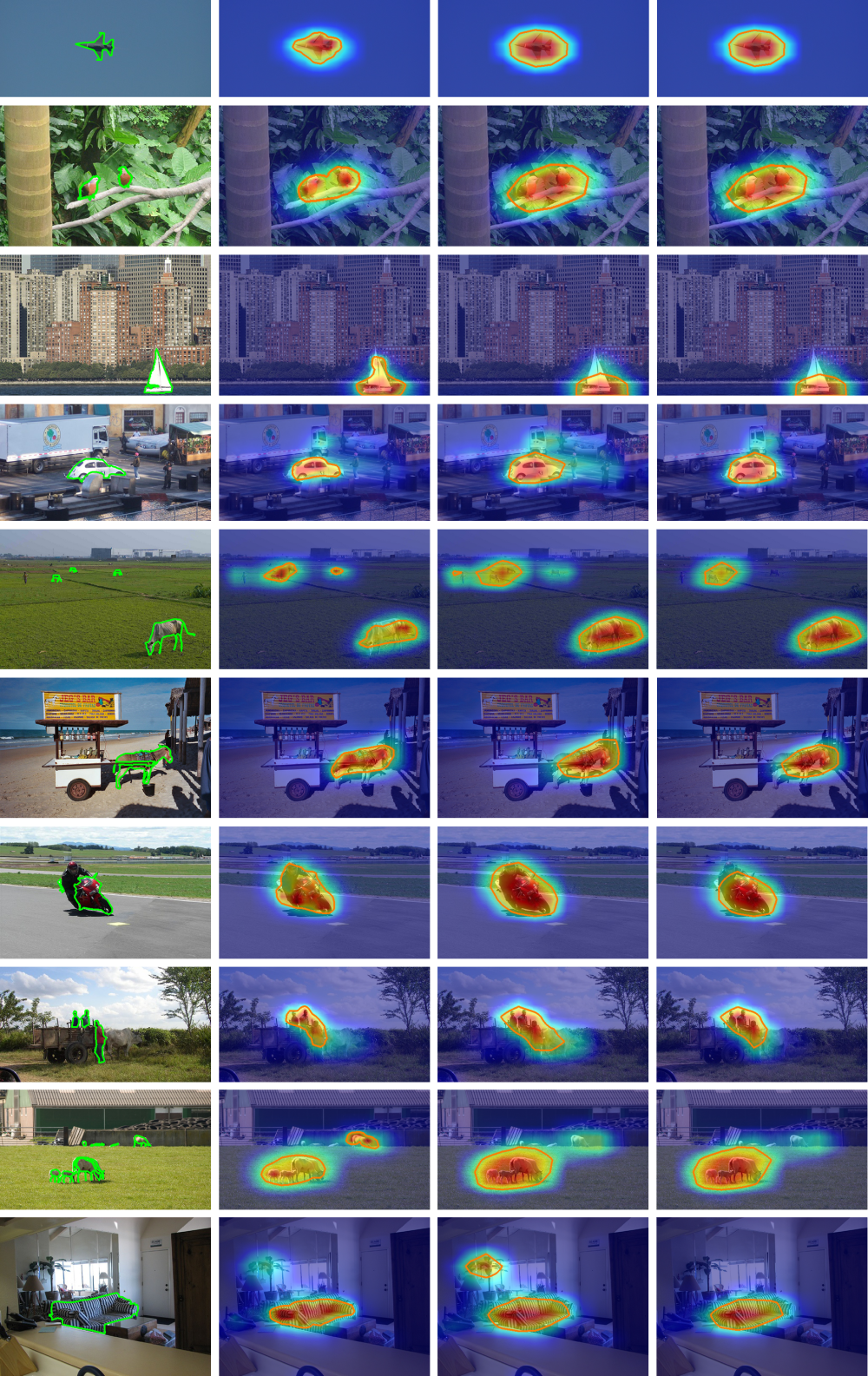}
   
    \put(2.5, 101){\scriptsize{Ground-truth}}
    \put(21.5,101){\scriptsize{\ourmethod{}}}
    \put(37.5,101){\scriptsize{\gcp{}}}
     \put(53.5,101){\scriptsize{\gc{}}}
    \end{overpic}
    \vspace{-2mm}
    \caption{Segmentation results based on saliency maps produced by GAM, GC, and GC++ (TResNet) on examples from Pascal VOC (validation) dataset, w.r.t. labels (top to bottom): \emph{aeroplane}, \emph{bird}, \emph{boat}, \emph{car}, \emph{cow}, \emph{horse}, \emph{motorbike}, \emph{person}, \emph{sheep} and \emph{sofa}.}
    \label{fig:loc_voc}
\end{figure}

\begin{figure}
    \vspace{10pt}
    \begin{overpic}[width=\columnwidth]{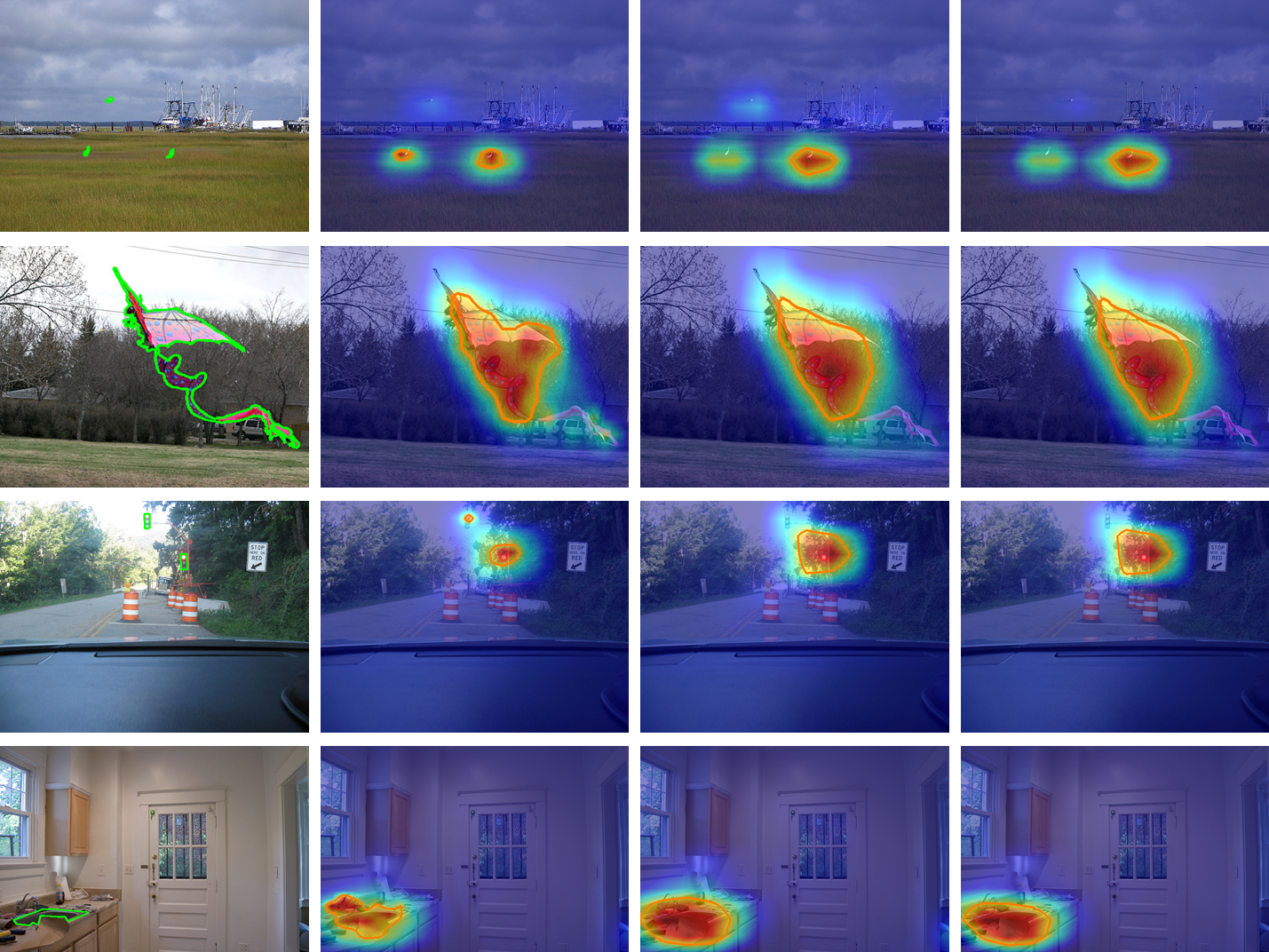}
    
    \put(4, 76){\scriptsize{Ground-truth}}
    \put(33.5,76){\scriptsize{\ourmethod{}}}
    \put(58.5,76){\scriptsize{\gcp{}}}
     \put(85,76){\scriptsize{\gc{}}}
    \end{overpic}
    \caption{Segmentation results based on saliency maps produced by GAM, GC, and GC++ (TResNet) on examples from MS-COCO (validation) dataset, w.r.t. labels (top to bottom): \emph{bird}, \emph{kite}, \emph{boat}, \emph{traffic light} and \emph{sink}.}
    \label{fig:loc_coco}
\end{figure}

\textbf{Localization of Small Objects:}
\label{subsec:detection_small}
We used the 25\% and 10\% partitions from Sec.~\ref{subsec:recognition} for testing the localization capability GAM on small objects. In addition, we conducted a localization experiment on medical imaging dataset ChestX-ray8 \cite{wang2017chestx}, where the classification decisions are usually made due to \emph{small} details in the images. In this experiment, we used the CheXNet model from \cite{rajpurkar2017chexnet} that was trained on the ChestX-ray8 dataset to classify common thorax diseases. 
The results for the ILSVRC-15-val 25\% / 10\% subsets and the ChestX-ray8 appear in the classification and similarity sections in Tab.~\ref{tab:loc}, and demonstrated in Fig.~\ref{fig:loc} (rows 3-4). We see that \ourmethod{} significantly outperforms both GC and GC++. These findings support the observation from Sec.~\ref{subsec:ablation} that multi-layer GAM ($n>1$) produces better saliency maps for small objects.

\textbf{Object Segmentation:}
\label{subsec:segmentation}
Finally, we tested the utilization of GAM, GC and GC++ for object segmentation. To this end, we applied the methods on top of two pretrained multi-label classification TResNet \cite{benbaruch2020asymmetric} models, trained on MS-COCO \cite{lin2014microsoft} and Pascal VOC \cite{everingham2010pascal} datasets. For each image, we computed the saliency maps w.r.t. each of the ground truth labels. Then, we computed the IoU\% of the binarized saliency map w.r.t. ground truth segmentation (in pixels), for each  ground truth label. The results appears in Tab.~\ref{tab:loc} (Segmentation), and exemplified in Fig.~\ref{fig:loc} (rows 5-6), and in Figs.~\ref{fig:loc_voc} and \ref{fig:loc_coco}. Overall, we see that GAM produces the most accurate segmentation.

\section{Conclusion}
\label{sec:Conclusion}
This work joins a growing effort to make machine learning models more transparent and explainable. To this end, we present GAM, a state-of-the-art method for explaining visual similarity and classification models in a unified manner. Extensive subjective and objective evaluations show that GAM outperforms its alternatives across various tasks and datasets, and especially on small objects. 

\bibliographystyle{ACM-Reference-Format}

\bibliography{ref}

\end{document}